\documentclass[12pt, reqno]{amsart}
\usepackage{amsmath, amsthm, amscd, amsfonts, amssymb, graphicx, color}
\usepackage[bookmarksnumbered, colorlinks, plainpages]{hyperref}
\usepackage{graphicx}
\usepackage[T1]{fontenc}
 \graphicspath{{./Pictures/}}
 
\usepackage{geometry}

\def\authorsaddresses#1{\dedicatory{#1}}
\numberwithin{equation}{section}
\usepackage{fancyhdr}
\begin{document}

\title[U-Net model for Cadastral Mapping]{Detecting Cadastral Boundary from Satellite Images Using U-Net model}

\author[N. Rahimpour]{Neda Rahimpour Anaraki}
\author[M. Tahmasbi]{Maryam Tahmasbi}
\author[SR. Kheradpishe]{Saeed Reza Kheradpisheh}
\authorsaddresses{Math science Department, Shahid Beheshti University, Tehran, Iran.\\
neda.rpa@gmail.com\\
m\textunderscore tahmasbi@sbu.ac.ir \\
s\textunderscore kheradpisheh@sbu.ac.ir\\}


\begin{abstract}

Finding the cadastral boundaries of farmlands is a crucial concern for land administration. Therefore, using deep learning methods to expedite and simplify the extraction of cadastral boundaries from satellite and unmanned aerial vehicle (UAV) images is critical. In this paper, we employ transfer learning to train a U-Net model with a ResNet34 backbone to detect cadastral boundaries through three-class semantic segmentation: "boundary", "field" and "background". We evaluate the performance on two satellite images from farmlands in Iran using "precision", "recall" and "F-score", achieving high values of 88\%, 75\%, and 81\%, respectively, which indicate promising results.

\end{abstract}

\maketitle

\section{Introduction}
Cadastral mapping, which records land ownership and physical locations, offers landowners security, sustainable livelihoods, and enhanced financial opportunities \cite{crommelinck2019application}. This makes cadastral mapping a significant focus in land administration. The importance is highlighted by the fact that around 75\% of land ownership is not registered in any official cadastral system, presenting a major challenge for both developed and developing countries \cite{luo2017quantifying, enemark2014fit}.

An effective cadastral system plays a crucial role in controlling unauthorized constructions, land market development and monitoring, property tax collection, urban infrastructure development, urban planning and generating statistical data, thereby enhancing land security \cite{crommelinck2016review}.

Maintaining up-to-date cadastral information is a high priority in countries that have spent decades creating comprehensive cadastral maps using traditional surveying methods \cite{kocur2021coherence}. However, in countries like Iran, creating cadastral maps is a significant challenge due to the time-consuming, costly, and labor-intensive nature of traditional methods \cite{koeva2020innovative}. Nowadays, with the availability of UAV images in some areas and globally observing satellites, remote sensing is used for cadastral mapping instead of field surveying. This approach, supported by land administration, speeds up operations and reduces costs \cite{enemark2014fit}.

Identifying cadastral boundaries from remote sensing images also presents challenges. Only boundaries that align with natural or man-made features, known as visible boundaries, can be identified. Therefore, cadastral mapping based on images recognizes that many cadastral boundaries in images are visible boundaries. Examples include roads, tree clusters, texture pattern changes, stone walls, fences, building walls, ditches, rivers, drainage channels, and strips of uncultivated land. These boundaries have the potential to be automatically extracted by image processing algorithms \cite{luo2017investigating, xia2019deep}.

Drăguţ et al. \cite{druaguct2014automated} introduced Multi-Resolution Segmentation (MRS) for multi-scale image segmentation using local variance to detect scale transitions. Classical edge detection identifies sharp changes in image brightness using first and second-order derivative-based methods.

Crommelinck et al. \cite{crommelinck2017contour} applied computer vision to UAV images for cadastral mapping through a three-step process: image preprocessing, boundary delineation using the Globalized Probability of Boundary (gPb) method, and image post-processing to create unified contour and binary boundary maps.

Crommelinck et al. \cite{crommelinck2019application} also used CNN tools in a three-step workflow: image segmentation to extract object outlines, boundary classification to predict boundary likelihood, and interactive delineation to connect lines based on predicted boundaries. For image segmentation, Multiresolution Combinatorial Grouping (MCG) generates object outlines, followed by boundary classification using Random Forest and pre-trained VGG19. Interactive delineation in QGIS creates final cadastral boundaries from UAV images.

Fetai et al. \cite{fetai2019extraction} used UAV images in a workflow of image pre-processing, boundary detection and extraction, and data post-processing. This involved resampling UAV orthoimages and applying the ENVI feature extraction module, followed by filtering and simplifying extracted objects 

Xia et al. \cite{xia2019deep} used deep Fully Convolutional Networks (FCNs) for detecting cadastral boundaries in UAV images. They treated boundary detection as a supervised pixel-wise classification task, using a modified FCN with dilated kernels. Also there are more studies in \cite{xu2023multiscale, cai2023improving, metaferia2023furthering}.

In this paper initially, we detail the 13 training images, 2 test images, and the process of creating masks for each training image. Four different filters and three buffer sizes were applied to each image and mask, respectively. We selected the best combination based on the highest "precision", "recall", and "F-score" values. Subsequently, we explore our approach to identifying the cadastral boundaries of farmlands. We utilize a U-Net model \cite{ronneberger2015u} with a ResNet34 backbone \cite{he2016deep}, pre-trained on the ImageNet dataset, to address the problem through three-class semantic segmentation: "boundary", "field" and "background". We train our model using four different configurations and selected those that yielded the best results based on the aforementioned metrics. In the post-processing section, we generate shapefiles of the output using skeleton \cite{zhang1984fast}, which are lines in vector format. Additionally, we convert each buffered boundary to a 1-pixel boundary. Then we evaluate assessment outputs of two test images by mentioned metrics and compare them to our previous results by Mask R-CNN model. In conclusion, we present our final findings.

\section{Data}

We used 13 images for training, of which 11 are satellite images obtained from Google Earth, and 2 are UAV images. All images, except for the one named "Ortho", are from farmlands in Iran, while "Ortho" is from Ethiopia. To prepare the images for the network, each image was divided into patches of 256 x 256 pixels. Details of each image can be found in Table \ref{tab:trimg}.
 
\begin{table}[h]
\centering
\caption{Details of train images.}\label{tab:trimg}
\begin{tabular}{|c|c|c|c|c|}
 \hline
 Name & Resolution & Image Type & Patches  \\
 \hline
 Aerial & 4963 x 2819 & UAV & 219  \\
  \hline
 Ortho & 3999 x 3999 & UAV & 239  \\
  \hline
 Famenin & 5520 x 3776 & Google Earth & 329  \\
  \hline
 Image1 & 4320 x 4160 & Google Earth & 272  \\
   \hline
 Image2 & 4096 x 4057 & Google Earth & 256  \\
    \hline
 Image4 & 4096 x 4077 & Google Earth & 256  \\
     \hline
 Image5 & 4096 x 4077 & Google Earth & 256  \\
     \hline
 Image7 & 4096 x 4067 & Google Earth & 256  \\
     \hline
 Image8 & 4096 x 4074 & Google Earth & 256  \\
     \hline
 Image10 & 5470 x 4160 & Google Earth & 351  \\
      \hline
 Image11 & 4096 x 4074 & Google Earth & 256  \\
      \hline
 Image12 & 4096 x 4074 & Google Earth & 256  \\
      \hline
 Image13 & 4858 x 2948 & Google Earth & 209  \\
\hline 
\end{tabular}
\end{table}

\begin{figure}[h]
\centering
\includegraphics[scale=0.6]{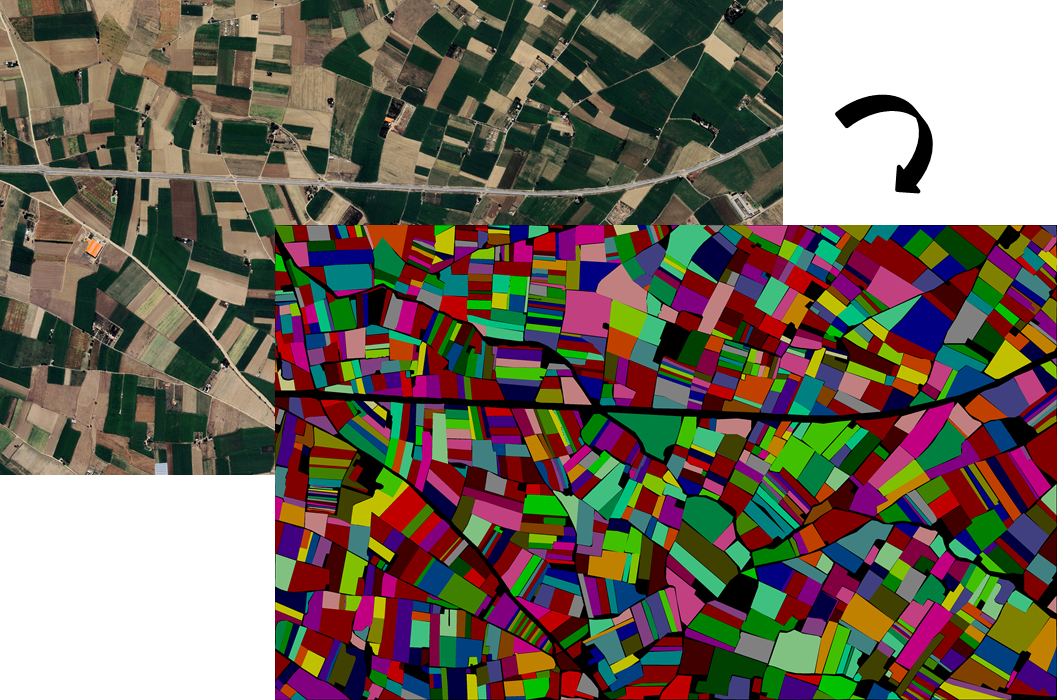}
\caption{Image13 and its corresponding initial mask.} \label{fig:img&mask}
\end{figure}

Since we are performing semantic segmentation and no public dataset meets our needs, we had to create corresponding masks for each training image by ourselves. Therefore, each mask was created using a software called LabelMe, which is a free graphical image annotation tool written in Python and uses Qt for its graphical interface \cite{wada2018labelme}. Each initial mask contains instance segmentation of farmlands, meaning that pixels belonging to each field are assigned individual colors. Fields that are not adjacent may share the same colors. In Figure \ref{fig:img&mask}, you can see one of the training images along with its corresponding mask, created as mentioned.

\section{Pre-Process}
\begin{figure}[h]
\centering
\includegraphics[scale=0.35]{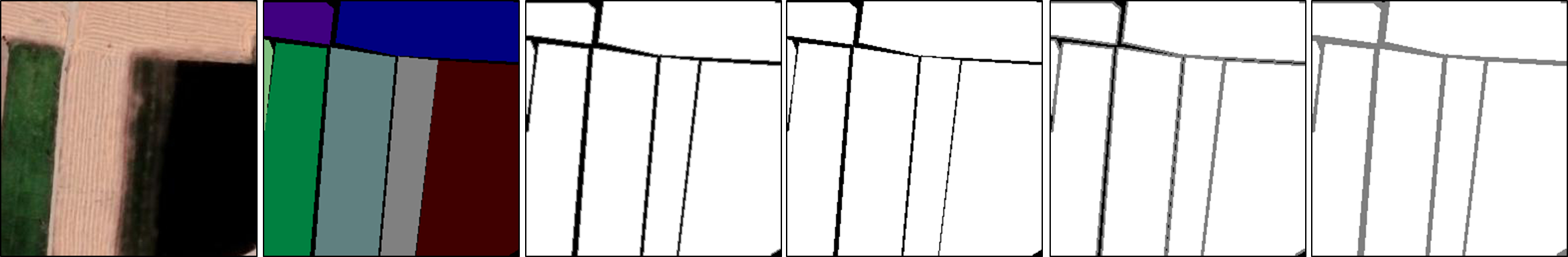}
\caption{From left to right, Original patch, it’s corresponding initial mask, secondary mask and final mask with buffer 1, 2 and 5 pixels.} \label{fig:pre1}
\end{figure}

After creating the initial mask, which is an RGB image, we eroded each field by 1 pixel to ensure that when all pixels belonging to the fields are turned white, no two neighboring fields merge into one. At this stage, all background pixels were turned black, creating a secondary mask. Finally, we added a gray boundary around each individual field. This process resulted in our final mask, which represents semantic segmentation across three classes: background, field and boundary (Figure \ref{fig:pre1}).

We also examined four different pre-processing techniques on each image in addition to the original image: High-pass filter, Laplacian filter, Sharp filter, and Sharp filter followed by Laplacian filter. 

\begin{itemize}
    \item High-pass filter: Isolates and emphasizes high-frequency components of the image, such as edges and fine details which Results in an image that is mostly black with highlighted edges and fine details.
    \item Sharp filter: Enhances overall image sharpness by increasing the contrast around edges and fine details. This filter maintains the original image appearance while enhancing the details, giving a crisper and more defined look.
    \item Laplacian filter: Specifically designed for edge detection by highlighting regions of rapid intensity change. It produces a grayish image with prominently displayed edges, showing where intensity changes occur most rapidly.
\end{itemize}

\begin{figure}[h]
\centering
\includegraphics[scale=0.5]{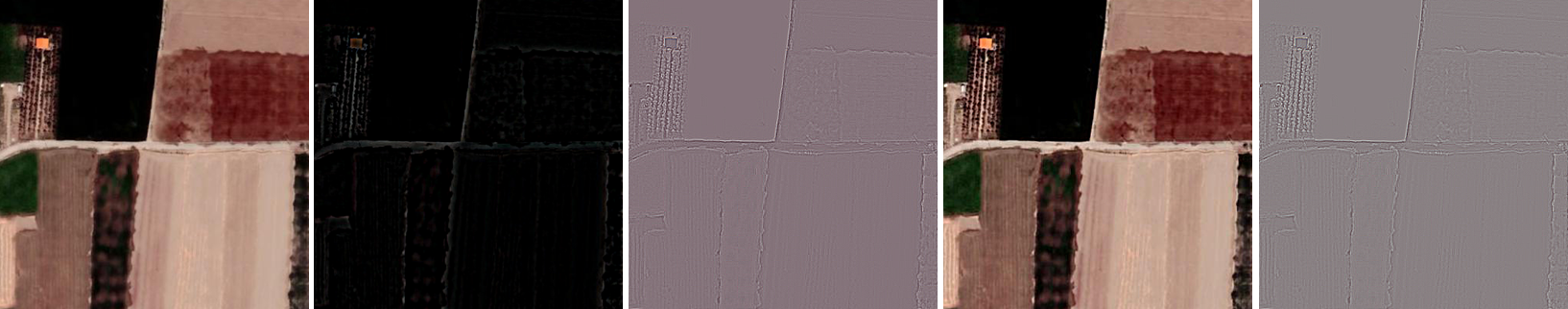}
\caption{From left to right, Original patch, high-pass filter, Laplacian filter, Sharped filter and Sharped then Laplacian filter applied on it.} \label{fig:pre2}
\end{figure}

 Results based on the three previous metrics indicated that applying the Laplacian filter to both the training and test images produced significantly better outputs than the other methods. In Figure \ref{fig:pre2}, the original patch is displayed alongside each of the four pre-processing techniques applied to it.

The two images chosen for testing are both satellite images taken from Google Earth, named “NegarKhatun” and “Karkhaneh,” both from Iran. The NegarKhatun image is from the farmlands of the village of the same name in Kangavar city, Kermanshah province, with a Ground Sample Distance (GSD) of 0.72 m. The Karkhaneh image, with a GSD of 0.56 m, is from the farmlands of the village of the same name in Famenin city, Hamedan province. NegarKhatun primarily consists of smallholder farms, while Karkhaneh has larger farmlands, with some patches entirely within a single field. Neither of these two images was used to train the network. The filters applied to each test image were identical to those used on the training images.

\section{Method}

\begin{figure}[h]
\centering
\includegraphics[scale=0.25]{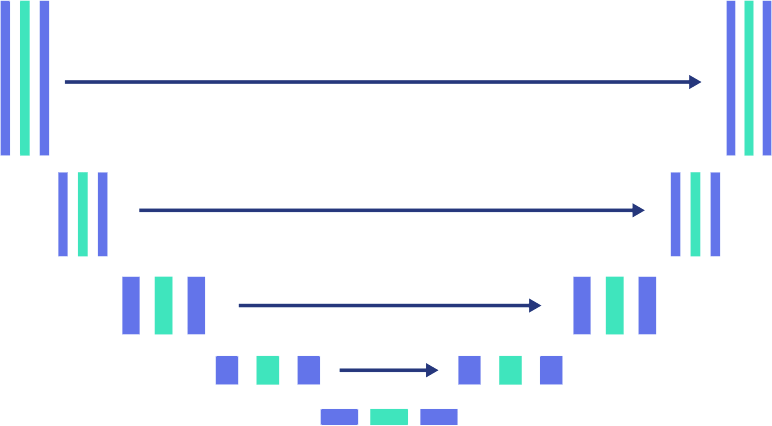}
\caption{U-Net model architecture.} \label{fig:unet}
\end{figure}

Our solution for finding the cadastral boundaries of farmlands involves using three-class semantic segmentation. Pixels belonging to the classes "field", "boundary" and "background" are represented as "white", "gray" and "black" respectively. Due to the limited number of training images, we utilized transfer learning. The proposed network is a U-Net model-inspired architecture (Figure \ref{fig:unet}) that leverages transfer learning by incorporating a pre-trained ResNet-34 as the encoder backbone. The network consists of two main components: an encoder for hierarchical feature extraction and a decoder for reconstructing high-resolution segmentation maps. This design ensures a balance between learning semantic features and preserving spatial resolution via skip connections.

The encoder is based on the pre-trained ResNet-34, which introduces residual blocks and batch normalization for enhanced feature learning and training stability. It operates in four stages, progressively downsampling the input resolution while doubling the number of feature channels:

\begin{itemize}
    \item Stage 1: The first convolutional layer uses a 7x7 kernel with 64 filters and a stride of 2. This is followed by batch normalization and ReLU activation.
    \item Stages 2–4: Each stage consists of a series of residual blocks, where the number of filters doubles at each stage: 64, 128, 256, and 512 filters, respectively. Max pooling or stride-2 convolutions are used for downsampling.
\end{itemize}

The final encoder output (bottleneck) consists of 512 filters, which are passed to the decoder.

The decoder mirrors the encoder with four stages of upsampling and convolution. At each stage, the decoder concatenates the upsampled feature maps with corresponding high-resolution feature maps from the encoder (skip connections), ensuring the preservation of spatial details. The number of filters decreases at each stage:

\begin{itemize}
    \item Upsampling Layers: UpSampling2D is used to double the spatial resolution of the feature maps.
    \item Convolutional Layers: Each stage applies two 3x3 convolutional layers with ReLU activation and batch normalization to refine the feature maps.
    \item The filter progression in the decoder is as follows: 512 → 256 → 128 → 64 → 32.
\end{itemize}

The last decoder stage outputs a segmentation map with the desired spatial resolution and 3 channels, corresponding to the number of target classes. This is achieved through a 1x1 convolution followed by a softmax activation function for multi-class segmentation.

Therefore, in total the encoder has 34 layers derived from ResNet-34, including residual blocks and the decoder has 8 convolutional layers, arranged in pairs across four stages, along with upsampling operations. The total number of trainable parameters is approximately 24.4 million, ensuring the network's capacity to handle complex segmentation tasks. ReLU activation is used throughout the encoder and decoder for non-linearity and softmax activation is applied in the final layer to produce normalized probabilities for each class.

Before choosing ResNet34 as the final backbone for the U-Net model, we trained the network with a ResNet50 backbone. ResNet50 has more layers and parameters than ResNet34, which can lead to overfitting, especially if the dataset is small or not very diverse. ResNet34, being shallower, generalizes better in this situation. The shallower architecture may capture the necessary spatial information without creating too much noise from deeper layers. Therefore, the additional capacity of ResNet50 may not be utilized effectively. ResNet architectures use skip connections to help with gradient flow. In our case, the shorter skip connections in ResNet34 facilitate more effective learning for certain tasks compared to the deeper connections in ResNet50.

The input images consisted of 11 satellite images from Google Earth and 2 UAV images. To feed them to the network, we experimented with two different patch sizes: 400 x 400 and 256 x 256 pixels. Patch size has a critical impact on the network’s output. When we used a 400 x 400 pixels patch size, the network failed to recognize many farmlands, detecting them as background pixels. Changing to a 256 x 256 pixels patch size improved the results. Therefore, creating a dataset of 256 x 256 pixels patches from each of the 13 training images resulted in 3358 training patches, with 20\% used as a validation set, yielding 2686 training patches and 672 validation patches.

We trained the U-Net model in four different ways, each consisting of various configurations: First, we used three types of combined cost functions. Second, we applied three buffer sizes to the boundary. Third, we considered four different pre-processing techniques for both training and test data, in addition to feeding the original patches to the network. Fourth, we experimented with different batch sizes and epochs. The following sections detail each approach.

First: We experimented with combinations of Tversky loss, Dice loss, and Jaccard loss with Focal loss. The first parts were used for multiclass segmentation, and the second part addressed class imbalance. The initial configurations of the network showed better results with the combination of Jaccard loss and Focal loss, leading us to choose this combination as our final cost function.

Second: We trained the network with three different buffer sizes for the boundary: 1, 2, and 5 pixels. The 1-pixel buffer caused the network to classify a large portion of pixels as "field". Since the 2-pixel buffer yielded better results, all subsequent results are based on this buffer. This means the detected boundary in the network’s output also has a 2-pixel buffer.

Third: As explained in the "Data" section of this paper, in addition to the original patches, we applied four different filters to each of them: High-pass filter, Laplacian filter, Sharp filter, and Sharp filter followed by Laplacian filter. Based on the superior results of the Laplacian filter, we selected this filter.

Fourth: To train the model, we fixed the learning rate at 0.001 and chose Adam as the optimizer. We experimented with batch sizes of 16 and 32, each for 100 and 200 epochs. The best output was achieved with a batch size of 32 for 200 epochs.

\begin{figure}[h]
\centering
\includegraphics[scale=0.65]{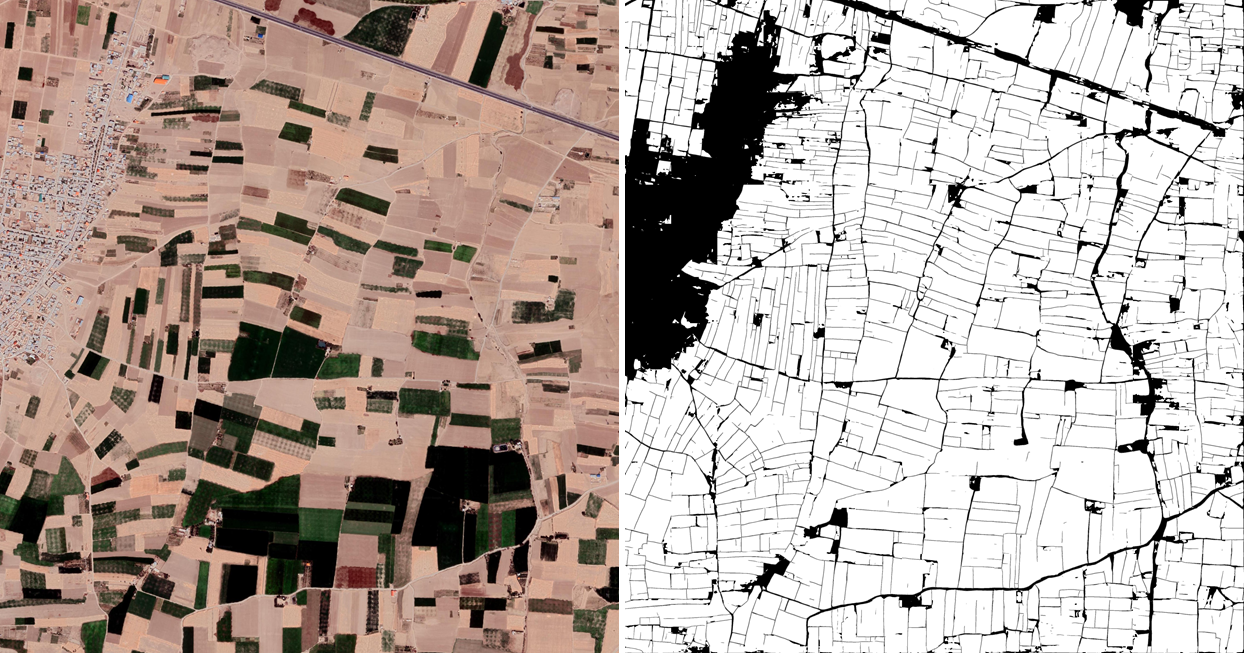}
\caption{NegarKhatun image and it’s corresponding initial output.} \label{fig:outn}
\end{figure}

\begin{figure}[h]
\centering
\includegraphics[scale=0.57]{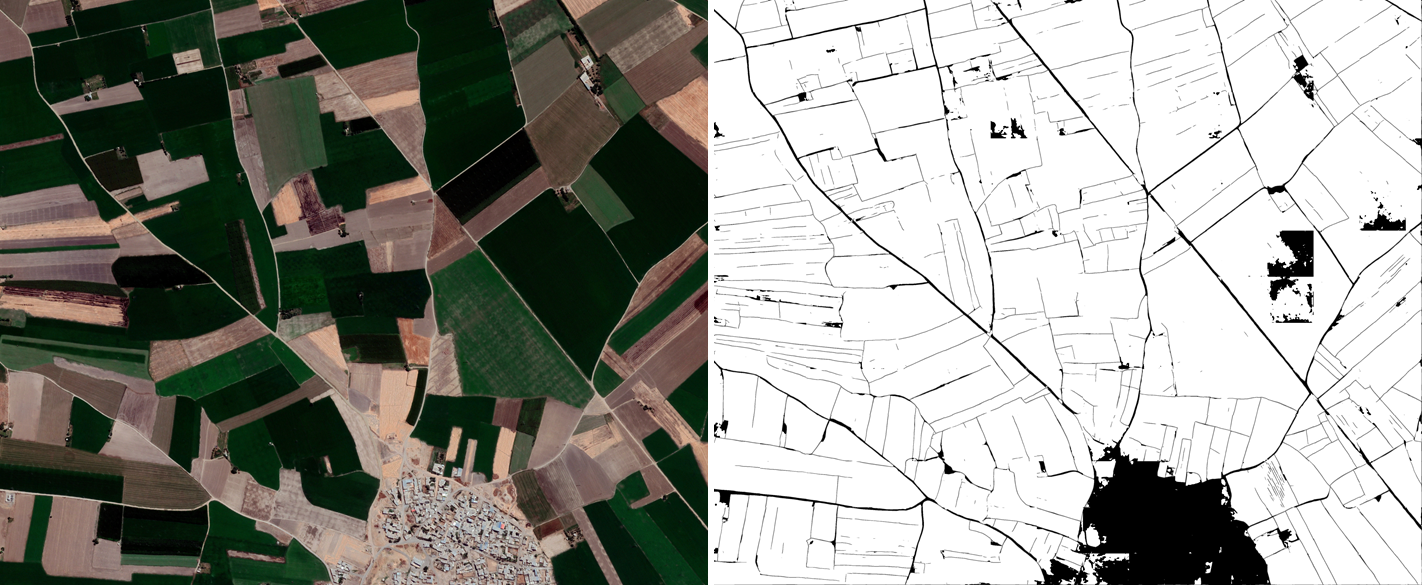}
\caption{Karkhaneh image and it’s correspoding initial output.} \label{fig:outk}
\end{figure}

The initial output of the U-Net model for each test image is illustrated in Figures \ref{fig:outn} and \ref{fig:outk}. This output represents three-class semantic segmentation, but we are only interested in two of these classes: boundary and background. Therefore, we performed some post-processing to obtain the final output.

\section{Post-Process}

To achieve the final output, we applied three post-processing techniques. First, as mentioned earlier, the width of each boundary in the U-Net model’s output is 2 pixels, but we need a 1-pixel boundary for evaluation. Second, we only need the "boundary" and "background" classes. Third, to work with the output in Geographic Information Systems (GIS), we need to convert the raster format of the boundary to vector format and save it as a Shapefile.

We achieved these goals by applying skeleton to the extracted boundary with a 2-pixel buffer, resulting in extracted lines (the skeleton of 2-pixel lines) that can be saved as both a vector in a shapefile and as a PNG image, which produces a 1-pixel boundary in raster format. Consequently, we have both raster and vector formats as our final output.

The skeleton is derived from the concept of the medial axis transform (MAT), a method for representing a shape by its skeleton. The MAT is generated by identifying points that have more than one closest point on the object's boundary, forming a network of lines or curves that capture the topology and geometry of the shape \cite{lee1994two}.

The extracted lines, i.e., boundaries, are clean enough—meaning they closely match their real structure and are not affected by zigzagging—that we see no need for further post-processing, such as simplifying them with the Douglas-Peucker algorithm.

\section{Evaluation}
We use three metrics—precision, recall, and F-score—to evaluate the results. precision indicates the percentage of valid boundaries among all predicted boundaries. recall represents the percentage of detected boundaries out of all reference boundaries. The F-score is the harmonic mean of precision and recall, serving as an overall assessment metric. 

Here, we use a confusion matrix for binary classification, where “boundary pixels” are the positive class and “non-boundary pixels” are the negative class. Formula for each metric is:

\begin{equation} \label{formula:precision}
precision =  \frac{TP}{TP+FP}
\end{equation} 

\begin{equation} \label{formula:recall}
recall =  BF \times \frac{TP}{TP+FN}
\end{equation} 

\begin{equation} \label{formula:fscore}
F-score =   \frac{2 \times precision \times recall}{precision+recall}
\end{equation} \\

As seen in the recall formula, there is an extra part called "BF". In this field, we always consider a certain buffer (BF) for the boundary, which is 2.4 m for rural areas and 0.3 m for urban areas \cite{IAAO2015}. Therefore, for the NegarKhatun image with a GSD of 0.72 m, we use a 5 and 6-pixel buffer (180 and 216 cm buffer, respectively), and for the Karkhaneh image with a GSD of 0.65 m, we use a 7 and 8-pixel buffer (196 and 224 cm buffer, respectively) for the reference boundary. It is worth mentioning again that no buffer is considered for the detected boundary. The reason for considering a buffer for the reference boundary is that TP + FP represents the total number of detected boundary pixels, while TP + FN indicates the total boundaries in the buffered reference, not the original reference. Therefore, to obtain the total boundaries in the original reference, which has a width of one pixel, we divide the sum of TP and FN by BF.

We compare the results obtained by the U-Net model in this paper with the results of our previous model, Mask R-CNN model \cite{rahimpour2024maskrcnn}, separately for each test image of NegarKhatun and Karkhaneh.

\begin{table}[h]
\centering
\caption{Accuracy Comparison of the Negarkhatun image.}\label{tab:res1}
\begin{tabular}{|c|c|c|c|c|c|}
 \hline
 Image & Model & Buffer & precision & recall & F-score  \\
 \hline
 NegarKhatun & Mask R-CNN & 5 pixels & 67 & 87 & 70  \\
  \hline
 NegarKhatun & Mask R-CNN & 6 pixels & 72 & 95 & 82    \\
  \hline
 NegarKhatun & U-Net & 5 pixels & 75 & 70 & 73  \\
 \hline
 NegarKhatun & U-Net & 6 pixels & 80 & 76 & 78 \\
\hline 
\end{tabular}
\end{table}

Table \ref{tab:res1} shows that U-Net model always achieves higher precision but lower recall than Mask R-CNN model, meaning U-Net model detects more valid boundaries than the other one. We can see that with a thiner buffer, U-Net model also achieves a higher F-score and is only beaten by Mask R-CNN model in the 6-pixel buffer due to its very high recall.

\begin{table}[h]
\centering
\caption{Accuracy Comparison of the Karkhaneh image.}\label{tab:res2}
\begin{tabular}{|c|c|c|c|c|c|}
 \hline
 Image & Model & Buffer & precision & recall & F-score  \\
 \hline
 Karkhaneh & Mask R-CNN	& 7 pixels & 71 & 85 & 75   \\
  \hline
 Karkhaneh & Mask R-CNN & 8 pixels & 74 & 91 & 81    \\
  \hline
 Karkhaneh & U-Net & 5 pixels & 82 & 66 & 73  \\
   \hline
 Karkhaneh & U-Net & 6 pixels & 85 & 69 & 76  \\
   \hline
 Karkhaneh & U-Net & 7 pixels & 87 & 72 & 79   \\
 \hline
 Karkhaneh & U-Net & 8 pixels  & 88 & 75 & 81 \\
\hline 
\end{tabular}
\end{table}

Higher precision for the U-Net model is also obtained in Karkhaneh image, as seen in Table \ref{tab:res2}. However, U-Net model achieves a higher F-score in all buffers, meaning U-Net model outperforms Mask R-CNN model in all stages. We also add the results of the 5-pixel buffer to the table to show the consistency of U-Net model. It shows a high percentage of detected boundaries lie within a 140 cm distance from the reference boundary, and only 8\% lie between 140 to 224 cm distance.

Overall, we can conclude that the Mask R-CNN model detects more boundaries but requires more correction due to its higher recall and lower precision. Conversely, the U-Net model detects more valid boundaries, although it doesn’t find as many boundaries as the Mask R-CNN model, based on its higher precision and lower recall. Additionally, there is a significant gap between the lower precision and higher recall of the Mask R-CNN model (23\% at best scenario), while in the U-Net model, this gap between higher precision and lower recall is smaller (13\% at best scenario). This strengthens the fact that the U-Net model needs less modification on detected boundaries and shows that most of the extracted boundaries are valid. Ultimately, the U-Net model outperforms the Mask R-CNN model in 5 out of 6 scenarios by achieving higher F-scores, making it a better option for finding cadastral boundaries of farmlands.

\begin{figure}[h]
\centering
\includegraphics[scale=0.48]{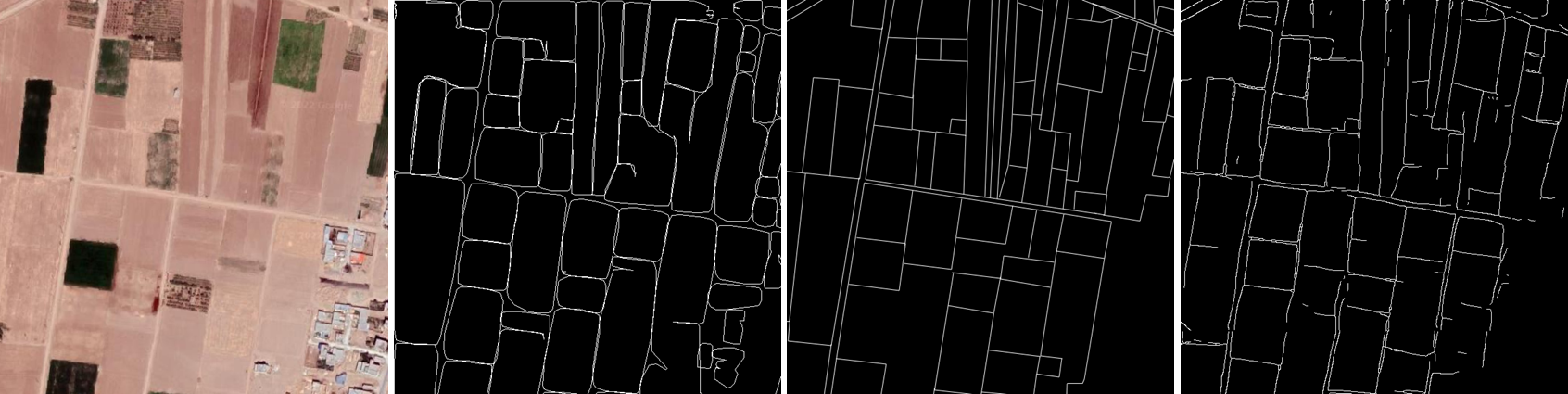}
\caption{A part of NegarKhatun image: From left to right. Original patch, Mask R-CNN model’s output, Reference Boundary and U-Net model’s output.} \label{fig:comp}
\end{figure}

When we examine the output of each model in the NegarKhatun and Karkhaneh images (Figure \ref{fig:comp}), we can see that the structure of each field extracted from the U-Net model is cleaner than that of the Mask R-CNN model, resulting in higher precision for the U-Net model. By "cleaner", we mean that fields detected by the Mask R-CNN model are rounded at the corners of their corresponding polygons and have serrated edges, which are improved only by geometric post-processing. However, even after applying geometric post-processing, the problem still exists and is only slightly improved from the initial output. On the other hand, the U-Net model’s fields have cleaner polygons even without any geometric post-processing. In fact, the Mask R-CNN achieves these results after applying three geometric post-processes (deleting small fields less than a real farmland, deleting fields that are wrongly detected inside another field, and simplifying over-segmentation), which has a significant impact on the initial output of the Mask R-CNN model.

It should be highlighted that the Mask R-CNN model failed to distinguish rural/urban areas from farmlands, meaning the final output also wrongly detected fields in rural/urban areas. Conversely, the U-Net model recognizes which parts of the image are rural-urban areas so well that it almost detects no wrong fields in those areas (Figure \ref{fig:comp}). This means the Mask R-CNN model requires pre-processing to remove rural/urban areas from the image first, and then the resulting image is given to the Mask R-CNN model. However, for the U-Net model, no pre-processing is needed as it detects rural/urban areas itself and avoids detecting wrong fields in these areas as farmland. Therefore, we can conclude that the U-Net model produces better visual results than the Mask R-CNN model too.

\section{Conclusion}
In this paper, we used a U-Net model with a ResNet34 backbone from the ImageNet dataset to perform three-class semantic segmentation on "boundary", "field" and "background". The input to the model can be an image with any size, which is divided into 256 x 256 patches. The output is a raster image where pixels belonging to the classes "boundary", "field" and "background" are colored gray, white, and black, respectively. However, since we are only interested in the "boundary" and "background" classes, we apply the skeleton method to the 2-pixel boundary in the raster format to extract a 1-pixel boundary in vector format, which is then saved as a shapefile. Thus, we provide both raster and vector formats as the final output.

Our final output is clean enough that there is no need for further processing, such as simplifying the extracted lines (i.e., boundaries). Additionally, our model can distinguish rural/urban areas from farmlands, so almost no fields are wrongly extracted in those areas. This means the model doesn't need pre- or post-processing to separate rural/urban areas.

Achieving high precision while maintaining high recall means the final output of our model requires fewer adjustments to align the detected boundaries with the reference boundaries. However, it should be noted that using the Skeleton method to extract lines and save them to a shapefile is not the best option. In some specific areas, it produces incorrect lines that were not part of the original boundary. Therefore, it is recommended to explore better methods for converting 2-pixel raster buffers into vector lines.

It is worth mentioning that, in our case, applying pre-processing filters to the original patches proved to be critical. These filters significantly affected the results, leading to promising outputs with high precision and reliable recall. Therefore, there may be other filters that could potentially improve the network's ability to produce even better outputs.


\bibliographystyle{Unsrt}
\bibliography{references}
\end{document}